\documentclass[sigconf]{acmart}
\graphicspath{{arxiv_figures/}}

\setcopyright{none}
\renewcommand\footnotetextcopyrightpermission[1]{}

\acmConference[]{}{}{}
\acmBooktitle{}
\acmISBN{}
\acmDOI{}
\acmPrice{}

\usepackage{booktabs}
\usepackage{multirow}
\usepackage{graphicx}
\usepackage{dsfont}

\begin{document}

\title{From Accuracy to Auditability: A Survey of Determinism in Financial AI Systems}

\settopmatter{authorsperrow=4} 
\author{Ruizhe Zhou}
\affiliation{\institution{Amazon.com}\country{USA}}
\email{rexzhou@amazon.com}

\author{Xiaoyang Liu}
\affiliation{\institution{Amazon.com}\country{USA}}
\email{lxaoya@amazon.com}

\author{Gaoyuan Du}
\affiliation{\institution{Amazon.com}\country{USA}}
\email{gdu@amazon.com}

\author{Yi Zheng}
\affiliation{\institution{Amazon.com}\country{USA}}
\email{yzhenmz@amazon.com}

\author{Shouxi Ren}
\affiliation{\institution{Amazon.com}\country{USA}}
\email{shouxi@amazon.com}

\author{Deepayan Chakrabarti}
\affiliation{\institution{Amazon.com}\country{USA}}
\email{deepayc@amazon.com}

\author{Dengdu Jiang}
\affiliation{\institution{Amazon.com}\country{USA}}
\email{dengdjiu@amazon.com}

\begin{abstract}
Deploying machine learning in regulated financial environments---credit risk, fraud detection, and anti-money laundering---exposes critical vulnerabilities in algorithmic reproducibility. While early financial ML addressed statistical challenges such as backtest overfitting, deep neural networks and Generative AI have introduced \emph{mechanical} nondeterminism rooted in hardware and architecture. This survey provides a systems perspective on reproducibility failures across three modalities now dominant in financial AI: tabular models (post-hoc explanation variance), graph networks (stochastic sampling and temporal asynchrony), and LLM-based agentic workflows (batch-dependent divergence and trajectory drift). We supplement the literature analysis with first-party experiments on public financial datasets---quantifying explanation rank instability in credit scoring, prediction flip rates in GNN-based fraud detection, and tensor-parallel-induced output divergence in LLM entity extraction. We propose a layered evaluation framework linking modality-specific metrics (RBO, $\mathcal{D}_{\text{cos}}$, TDI, PSD) to audit readiness, and report where these measures overlap rather than complement one another.
\smallskip
\noindent\emph{Note on this version (v3).} This version corrects several errors in
v1--v2 that were identified while re-running the same experiments for follow-on work.
The graph flip rates were computed on a single split while described as covering ten;
the cross-configuration LLM divergence rate was reported at half its true value because
pooled exact match has a floor at $0.5$; the three ``regimes'' claimed for the TDI--PSD
relationship are withdrawn as an artefact of a proxy implementation of PSD; the stated
mechanism for GCN's higher flip rate is contradicted by direct measurement; and
``$M{=}20$'' on German Credit denoted an undisclosed variance-selected subset of a
48-column encoding. Corrected values supersede v1--v2 wherever they differ and are
marked in place.
\end{abstract}

\maketitle
\thispagestyle{plain}
\pagestyle{plain}

\section{Introduction}

Deploying machine learning in financial risk management---Credit Scoring, Fraud Detection, and Anti-Money Laundering (AML)---requires not only predictive accuracy but also algorithmic reproducibility. ML models in these domains operate under strict mandates such as the US Federal Reserve's SR~11-7 guidelines, the Equal Credit Opportunity Act (ECOA), and the European Union's AI Act~\cite{bis2024mrm, eu2024aiact}. A model whose outputs vary materially across independent runs under identical inputs poses severe challenges for legal auditability, potentially rendering it undeployable regardless of its performance.

Throughout this survey, we distinguish three related concepts. \textit{Computational determinism} requires bit-exact identical outputs given identical inputs, model weights, and hardware---the strongest guarantee. \textit{Reproducibility} is the weaker requirement that independent executions yield outputs consistent enough to support the same downstream decision (e.g., the same credit denial reasons), even if surface-level tokens differ. \textit{Auditability} requires that any algorithmic decision can be reconstructed and independently verified after the fact, presupposing at least reproducibility.

Historically, nondeterminism in financial modeling was understood as a statistical phenomenon---selection bias, multiple testing, and distributional shift. Harvey et al.\ (2016) exposed the ``factor zoo'' of overfitted factors, while Bailey and L\'{o}pez de Prado (2014) formalized how selection bias inflates the Sharpe ratio~\cite{harvey2016cross, bailey2014deflated, arian2024backtest}. However, a fundamentally different class of nondeterminism has emerged---one rooted in the computational machinery itself. Modern ML architectures introduce nondeterminism through non-associative floating-point reductions on parallel accelerators, stochastic algorithmic components (GNN neighborhood sampling during inference, dropout during training), and emergent instabilities in agentic systems (LLM trajectory drift).

Even when researchers fix all random seeds and use greedy decoding, exact replication is thwarted by the hardware stack. Modern parallelism paradigms~\cite{narayanan2021megatron} compound this: during training, data parallelism introduces fluctuating \textit{All-Reduce} accumulation order and pipeline parallelism creates timing variances in gradient application; during inference, tensor parallelism injects numerical noise via \textit{All-Gather} collectives and dynamic batching alters reduction order within attention kernels. Propagated across billions of parameters, these divergences cascade through non-linear activations into macroscopic variations.

A significant translation gap persists between regulatory expectations and technical practice. The EU AI Act~\cite{eu2024aiact} articulates governance principles but relies on standards bodies to define acceptable variance. Standard metrics like AUROC or F1 are insensitive to run-to-run variance. Existing compliance approaches often append post-hoc XAI methods to satisfy interpretability requirements without resolving underlying mechanical instability.

This survey systematically examines nondeterminism across three modalities: tabular data (post-hoc explanation variance in credit scoring), dynamic graphs (stochastic sampling and temporal asynchrony in fraud detection), and unstructured text (hardware-level nondeterminism and trajectory drift in agentic AML workflows). We conducted a targeted literature search across Google Scholar and arXiv (2018--2025), prioritizing work reporting quantitative reproducibility measurements over qualitative discussion.

\paragraph{Contributions.}
First, we delineate classical statistical reproducibility concerns from modern systems-level computational nondeterminism. Second, we provide a source-verified taxonomy of nondeterminism mechanisms across three modalities, supplemented by first-party experiments on public financial datasets (Sections~2, 3, and~4). Third, we propose a layered, modality-aware evaluation framework---linking feature attribution consistency (RBO), graph embedding variance ($\mathcal{D}_{\text{cos}}$), and agentic output agreement (TDI, PSD) to regulatory audit readiness---bridging domain-agnostic reproducibility tools~\cite{detlefsen2022torchmetrics, bhojanapalli2021reproducibility} and financial compliance requirements.

\paragraph{Organization.}
Section~2 examines tabular explainability instability. Section~3 studies graph-based fraud detection. Section~4 analyzes LLM/agentic nondeterminism. Section~5 reviews engineering solutions. Section~6 proposes evaluation metrics. Section~7 concludes.

\section{Tabular Risk Models: The Determinism Challenge in Post-Hoc Explainability}

In the domains of credit and fraud risk assessment, the foundational data modality remains strictly tabular. Historically, financial institutions relied on highly interpretable, deterministic frameworks such as generalized linear models (GLMs) or logistic regression. However, the pursuit of higher predictive accuracy has driven the industry toward highly parameterized, non-linear architectures, specifically deep ensemble trees (e.g., XGBoost, LightGBM) and deep tabular neural networks (e.g., TabNet) \cite{arik2021tabnet}. While these models capture complex, non-monotonic feature interactions, they function as computational black boxes. This creates a direct conflict with regulatory mandates: lending laws such as the Equal Credit Opportunity Act (ECOA) require explicit, deterministic ``Adverse Action Notices,'' and anti-fraud operations require auditable Suspicious Activity Reports (SARs) \cite{bis2024mrm}. To bridge the gap between black-box accuracy and regulatory compliance, the ML community initially adopted post-hoc Explainable AI (XAI) frameworks, most notably LIME and SHAP \cite{lundberg2017unified, ribeiro2016should}. However, treating explainability as a post-hoc software patch over a probabilistic model may introduce reproducibility failures.

\subsection{The Challenge: Algorithmic Instability and Sampling Variance}

The fundamental determinism challenge in tabular risk models stems from the computational complexity of post-hoc attribution. The theoretical foundation of SHAP rests on cooperative game theory, requiring the evaluation of the model over the entire power set of features to compute exact marginal contributions. For a dataset with $M$ features, this necessitates $2^M$ model evaluations, rendering exact computation NP-hard. To circumvent this, practitioners rely on approximation algorithms, predominantly KernelSHAP, which samples binary feature coalitions from a background distribution \cite{lundberg2017unified}. This introduces a critical vulnerability: \textit{algorithmic instability driven by sampling variance}. As demonstrated by Covert and Lee (2021), the variance of the KernelSHAP estimator scales at $\mathcal{O}(1/n)$, where $n$ is the number of sampled coalitions \cite{covert2021improving}. Consequently, two independent executions of KernelSHAP on the exact same credit application, using the exact same underlying model but different random seeds, will yield divergent feature importance rankings.

The magnitude of this variance is severe. Covert and Lee report that on the German Credit dataset, standard KernelSHAP requires 7--13$\times$ more samples to converge to a stable mean-squared error threshold compared to paired-sampling variants, while unbiased estimators require up to 17{,}437$\times$ more samples. Direct financial evidence confirms this: Lin and Wang (2025) measured SHAP ranking stability across 100 XGBoost reruns on credit default data \cite{lin2025shapcredit}. While the top-ranked feature was stable, mid-ranked features---precisely the borderline variables most likely to appear in customer-facing denial rationales---spanned up to 25 distinct rank positions due to initialization and sampling variance. For an auditor, this means the legal reason provided for a credit denial is a nondeterministic statistical artifact.

\subsection{The Challenge: Explanation Drift and Adversarial Fragility}

Beyond internal sampling variance, post-hoc explainers exhibit extreme sensitivity to localized data geometries, resulting in explanation drift. Because algorithms like LIME and KernelSHAP operate by perturbing inputs to observe the black-box model's local behavior, the explainer's decision boundary is inherently fragile. Seminal research by Slack et al. (2020) proved that this sampling mechanism can be actively exploited \cite{slack2020fooling}. An adversary can train a biased ``scaffolding'' model that executes discriminatory lending policies on the actual data distribution but behaves benignly on the explainer's perturbed distribution. In their experiments, adversarial attacks displaced the true sensitive feature from the top explanation ranks in 85--100\% of held-out instances on financial datasets.

Even without malicious intervention, natural data drift triggers this fragility. A minor, imperceptible shift in a continuous feature (e.g., an applicant's income changing by a fraction of a percent) frequently leaves the model's final risk score unchanged but completely reshuffles the local perturbation neighborhood, radically altering the final SHAP feature hierarchy \cite{ghorbani2019interpretation}. This proves that local decision boundaries estimated by post-hoc tools lack the mathematical stability required for reliable Model Risk Management.

\subsection{Mitigations}

Three structural solutions enforce determinism at the explanation layer.

\textbf{TreeSHAP.} Lundberg et al.~\cite{lundberg2020local} developed TreeSHAP for tree ensembles (XGBoost, LightGBM), computing exact attributions in $\mathcal{O}(TLD^2)$ by pushing Shapley computation into the tree structure. This restores bit-exact determinism without stochastic approximation.

\textbf{Explainable Boosting Machines (EBMs).} EBMs~\cite{lou2012intelligible, caruana2015intelligible} abandon the black-box paradigm: each feature's contribution $f_j(x_j)$ is computed independently via isolated shallow trees, with the final prediction $g(E[y]) = \beta_0 + \sum_j f_j(x_j)$. No perturbation or sampling is required---explanations are read from a static lookup table.

\textbf{Stability-aware training.} Addressing findings that cost-sensitive optimization degrades explanation stability~\cite{ballegeer2025creditstability}, adversarial regularization penalizes large attribution shifts under minor input perturbations, smoothing local decision boundaries.

Table~\ref{tab:tabular_solutions} summarizes these solutions.

\begin{table}[t]
\centering
\caption{Determinism challenges and solutions in tabular risk modeling.}
\label{tab:tabular_solutions}
\small
\resizebox{\columnwidth}{!}{%
\begin{tabular}{p{2.2cm}p{2.8cm}p{4.5cm}}
\toprule
\textbf{Challenge} & \textbf{Source} & \textbf{Mitigation} \\
\midrule
Sampling Variance & NP-hard Shapley requires stochastic approx. & \textbf{TreeSHAP:} Exact polynomial-time structural routing \\
\addlinespace
Explanation Drift & Perturbation fragility & \textbf{Adversarial regularization:} Penalize high-gradient shifts \\
\addlinespace
Black-Box Obfuscation & Post-hoc patches & \textbf{EBMs:} Inherently interpretable, variance-free lookup tables \\
\bottomrule
\end{tabular}}
\end{table}

\subsection{Empirical validation.}
To provide first-party evidence for the claims above, we conduct a controlled rank-stability experiment on two public credit datasets: the UCI German Credit dataset ($N{=}1{,}000$) and the Kaggle Default of Credit Card Clients dataset ($M{=}23$, $N{=}30{,}000$).

\emph{A disclosure about $M$ that v1--v2 omitted.} German Credit has 20 original attributes, most of them categorical, so the number of \emph{model inputs} depends on the encoding. Versions v1--v2 reported ``$M{=}20$'' for a pipeline that one-hot encoded the categoricals into 48 columns and then reduced back to 20 by ranking columns on variance. That $M{=}20$ was therefore a variance-selected subset of the encoded space, neither the dataset's native attributes nor the full encoding, and the selection step was not disclosed. It also does not do what it appears to: a dummy's variance is $p(1-p)$, so ranking one-hot columns by variance ranks how near a category's frequency is to $50\%$ rather than how informative it is, and it makes $M$ a function of the encoding, which undercuts the explanation that larger $M$ enlarges the permutation space. We have removed the step and report \emph{both} encodings below, since the gap between them is a result rather than a nuisance: 20 integer-coded attributes, and the full 48-column one-hot space. For each dataset, we train XGBoost classifiers using 3 independent random seeds and evaluate explanation stability using two explainers: KernelSHAP (with sample budgets $n \in \{100, 500, 1{,}000, 5{,}000\}$) and TreeSHAP (exact computation). For each of 50 randomly sampled test instances, we execute each explainer configuration 30 times with distinct PRNG seeds, holding the underlying model weights fixed. We measure rank stability using the Jaccard Index at $k{=}3$ and $k{=}5$ (the number of adverse action reasons typically required by ECOA), and Rank-Biased Overlap (RBO) with decay parameter $p{=}0.9$.

Table~\ref{tab:shap_stability} reports the results, which supersede those in v1--v2 for the reason just given. TreeSHAP achieves perfect determinism ($J@3 = J@5 = \text{RBO} = 1.0$) across all instances and all three configurations, confirming that exact structural computation eliminates explanation variance entirely. KernelSHAP at the common default of $n{=}100$ does not: on German Credit at $M{=}20$, $J@3$ averages $0.74$, so roughly one run in four produces a different set of top-3 denial reasons for the same applicant. Increasing the budget to $n{=}5{,}000$ reaches $0.97$ and still falls short of the $J@3{=}1.0$ a binding disclosure requires.

Encoding width matters, and the within-dataset comparison isolates it: holding the data, the model family and the sampled instances fixed, moving German Credit from 20 integer-coded attributes to the 48-column one-hot space lowers $J@3$ at $n{=}100$ from $0.74$ to $0.61$ and $J@5$ from $0.71$ to $0.54$. The cross-dataset ordering points the same way (Credit Card Default, $M{=}23$, reaches $J@3{=}0.71$ and $J@5{=}0.62$ at $n{=}100$), but that comparison confounds $M$ with $N$ and with the data-generating process, so we do not rest the claim on it.

One further correction to the RBO column. The values in v1--v2 omitted the tail term of Rank-Biased Overlap, which accounts for the unobserved remainder of the ranking beyond depth $k$; without it, RBO is systematically understated and cannot reach $1.0$ even for an exactly deterministic explainer. The column below uses the complete definition, $\text{RBO} = (1-p)\sum_d p^{d-1} A_d + p^{k} A_k$, and is correspondingly higher throughout.

\begin{table}[t]
\centering
\caption{Explanation rank stability across 30 independent runs per instance (50 test
instances, 3 model seeds). Higher is better ($\uparrow$). German Credit is reported
under both encodings, since v1--v2's ``$M{=}20$'' was a variance-selected subset of the
48-column one-hot space rather than either encoding (Section~2.4). RBO uses the
complete definition including the tail term; v1--v2 omitted it and understated the
column throughout. These values supersede v1--v2.}
\label{tab:shap_stability}
\small
\resizebox{\columnwidth}{!}{%
\begin{tabular}{llccc}
\toprule
\textbf{Dataset} & \textbf{Explainer (samples)} & \textbf{J@3 $\uparrow$} & \textbf{J@5 $\uparrow$} & \textbf{RBO $\uparrow$} \\
\midrule
\multirow{5}{*}{German Credit ($M{=}20$)}
 & KernelSHAP ($n{=}100$) & $0.74$ & $0.71$ & $0.86$ \\
 & KernelSHAP ($n{=}500$) & $0.91$ & $0.89$ & $0.94$ \\
 & KernelSHAP ($n{=}1000$) & $0.92$ & $0.92$ & $0.96$ \\
 & KernelSHAP ($n{=}5000$) & $0.97$ & $0.97$ & $0.98$ \\
 & TreeSHAP (exact) & $1.00$ & $1.00$ & $1.00$ \\
\addlinespace
\multirow{5}{*}{German Credit ($M{=}48$, one-hot)}
 & KernelSHAP ($n{=}100$) & $0.61$ & $0.54$ & $0.69$ \\
 & KernelSHAP ($n{=}500$) & $0.85$ & $0.82$ & $0.87$ \\
 & KernelSHAP ($n{=}1000$) & $0.90$ & $0.87$ & $0.91$ \\
 & KernelSHAP ($n{=}5000$) & $0.95$ & $0.94$ & $0.96$ \\
 & TreeSHAP (exact) & $1.00$ & $1.00$ & $1.00$ \\
\addlinespace
\multirow{5}{*}{Credit Card Default ($M{=}23$)}
 & KernelSHAP ($n{=}100$) & $0.71$ & $0.62$ & $0.80$ \\
 & KernelSHAP ($n{=}500$) & $0.84$ & $0.81$ & $0.91$ \\
 & KernelSHAP ($n{=}1000$) & $0.88$ & $0.86$ & $0.93$ \\
 & KernelSHAP ($n{=}5000$) & $0.94$ & $0.94$ & $0.97$ \\
 & TreeSHAP (exact) & $1.00$ & $1.00$ & $1.00$ \\
\bottomrule
\end{tabular}}
\end{table}

\begin{figure}[t]
\centering
\includegraphics[width=\columnwidth]{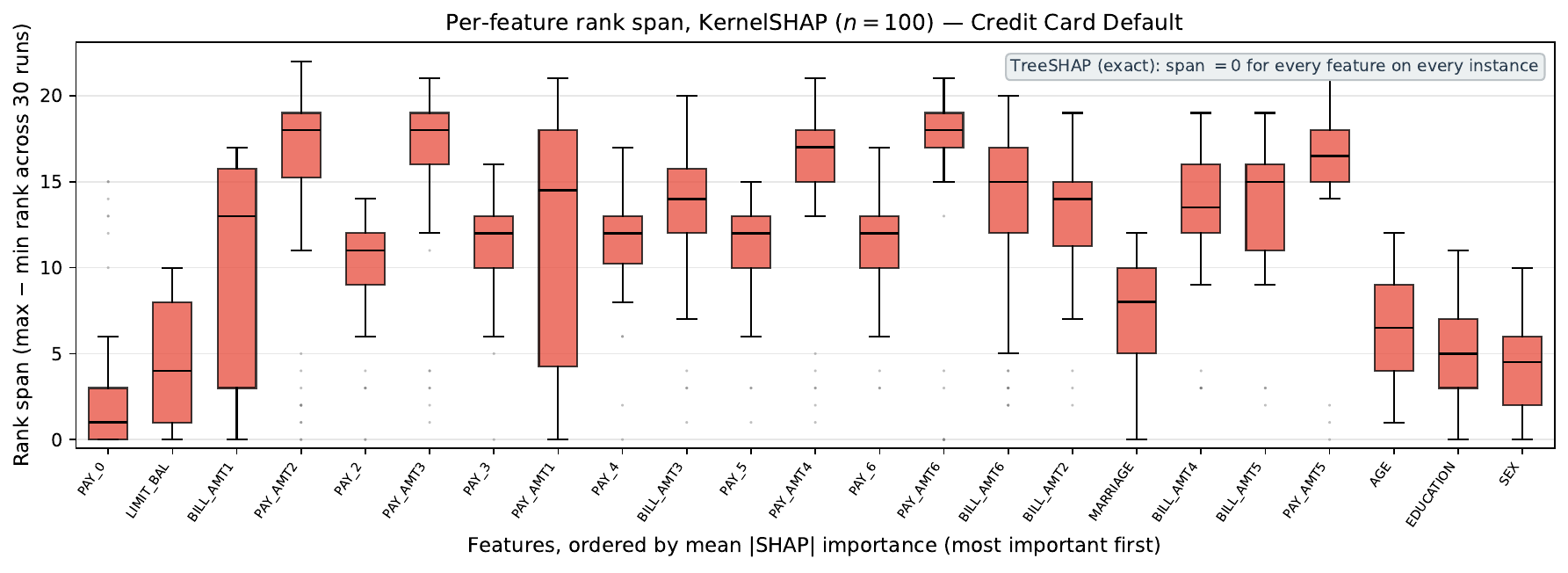}
\caption{Feature rank span under KernelSHAP ($n{=}100$) on Credit Card Default
($M{=}23$). Each box is the distribution, over the 50 sampled test instances, of the
range of rank positions that feature occupies across 30 independent runs. Features are
ordered left to right by mean $|$SHAP$|$ importance, so horizontal position is
importance and height is instability. Median span by importance tercile is $11.0$
(top), $14.3$ (middle), $10.4$ (bottom): the mid-ranked third, the features that fill
out an adverse-action notice beyond the first reason, is the least stable. TreeSHAP
yields zero span for every feature on every instance. \emph{The caption in v1--v2
reported ``up to 25 rank positions of variance'' here; with 23 features the maximum
possible span is 22, and that figure was Lin and Wang's (\S2.1) rather than ours.}}
\label{fig:rank_span}
\end{figure}

Figure~\ref{fig:rank_span} provides a complementary per-feature view: while Table~\ref{tab:shap_stability} measures top-$k$ set overlap, Figure~\ref{fig:rank_span} shows the rank span (max rank $-$ min rank across 30 runs) for each individual feature. Mid-ranked features---those most likely to appear in customer-facing denial rationales---exhibit the widest rank spans under KernelSHAP.

\section{Graph-Based Fraud Detection: The Mechanics of Topological Nondeterminism}

While tabular data dominates financial machine learning, modern fraud rings and Anti-Money Laundering (AML) typologies are inherently relational. Malicious actors exploit complex networks of synthetic identities, shell companies, and layered transactions to obscure illicit fund flows. To capture these topological dependencies, the financial industry has rapidly transitioned from isolated tabular models to Graph Neural Networks (GNNs) \cite{kipf2016semi}. For example, in these architectures, nodes might represent financial entities—such as user accounts, IP addresses, or routing numbers—while edges represent their interactions, like wire transfers or shared devices.

Unlike tabular ML, where a model processes an isolated feature vector, GNNs rely on recursive message-passing frameworks. A node's latent representation (embedding) is iteratively updated by aggregating the feature vectors of its localized topological neighborhood. However, the application of GNNs to web-scale financial transaction graphs introduces systems-level reproducibility challenges \cite{weber2024gnnfraud}. These failures originate not from statistical overfitting, but from the mechanical necessity of stochastic sampling, the asynchronous nature of dynamic graphs, and hardware-level atomic operations on the GPU.

\subsection{Stochastic Neighborhood Sampling}

Full-batch GNNs require the entire adjacency and feature matrices in GPU memory---infeasible for enterprise AML networks. GraphSAGE/PinSAGE~\cite{hamilton2017graphsage, ying2018pinsage} sample stochastic neighbor subsets, guaranteeing run-to-run variance~\cite{shchur2018pitfalls}. Shchur et al.\ show model rankings reverse across random splits (100 splits $\times$ 20 initializations): on Cora, GAT leads under Planetoid but GCN leads under random splits (see Table~\ref{tab:graph_instability} in Appendix~\ref{app:shchur}). Domain-specific work confirms this: Deprez et al.~\cite{deprez2024aml} find synthetic AML data yields ``overly optimistic results''; Duy et al.~\cite{duy2026elliptic} show initialization and normalization choices materially affect GNN performance on Elliptic Bitcoin.

\paragraph{Financial-domain validation.}
We evaluate on the Elliptic Bitcoin dataset (203K nodes, 234K edges, ${\sim}2\%$ illicit)~\cite{weber2024gnnfraud} with GCN under deterministic aggregation and GraphSAGE and GAT under default aggregation, each over 50 seeds $\times$ 10 splits (2500 training runs). The corrected figures below come from a re-run on Tesla T4s, one configuration per device; the superseded v1--v2 figures were produced on an A10G. \textbf{All three train full-batch:} we vary the aggregation operator and the determinism flag, not the sampling regime, and no neighbor sampling is exercised anywhere in this experiment. Earlier versions of this survey labelled the GraphSAGE and GAT rows with fan-out sizes, which corresponded to nothing in the code and are removed.

Table~\ref{tab:elliptic_f1} shows GraphSAGE attaining the highest F1 ($0.799$) while $15.5\%$ of nodes deviate from their modal label across seeds. Flip rates are computed within each split over its 50 seeds and then averaged across the 10 splits, since test node sets differ by split. \emph{The figures in v1--v2 ($22.8/30.0/28.5\%$) were computed on split~0 alone while described as covering ten splits, and are superseded by the values in Table~\ref{tab:elliptic_f1}.}

GCN's higher flip rate ($27.5\%$) despite deterministic execution is \emph{not} explained by weaker discrimination near the boundary, as v1--v2 asserted. Retaining per-node cross-seed scores, mechanical dispersion is \emph{highest} for GraphSAGE (mean s.d.\ $0.0867$), then GCN ($0.0758$), then GAT ($0.0683$)---the reverse of the flip-rate ordering on the same nodes (GCN $32.0\%$, GAT $27.7\%$, GraphSAGE $21.9\%$ on split~0). GraphSAGE's low flip rate is bought with margins and not with determinism: its scores move \emph{more} run to run than either competitor's. A flip rate mixes pipeline noise with distance to the decision threshold, so it is not an instrument for mechanical nondeterminism, and a tolerance written against it does not bound the serving stack.

Because the flip rate counts \emph{any} deviation from the modal label, it is monotone non-decreasing in the number of runs $R$; a figure quoted without its $R$ is not interpretable. The $R$-invariant companion is the expected pairwise disagreement (EPD), the probability that two independent runs disagree, which at $R{=}50$ is four to nine times smaller than the any-deviation form. Table~\ref{tab:elliptic_f1} reports both.

\begin{figure}[t]
\centering
\includegraphics[width=0.8\linewidth]{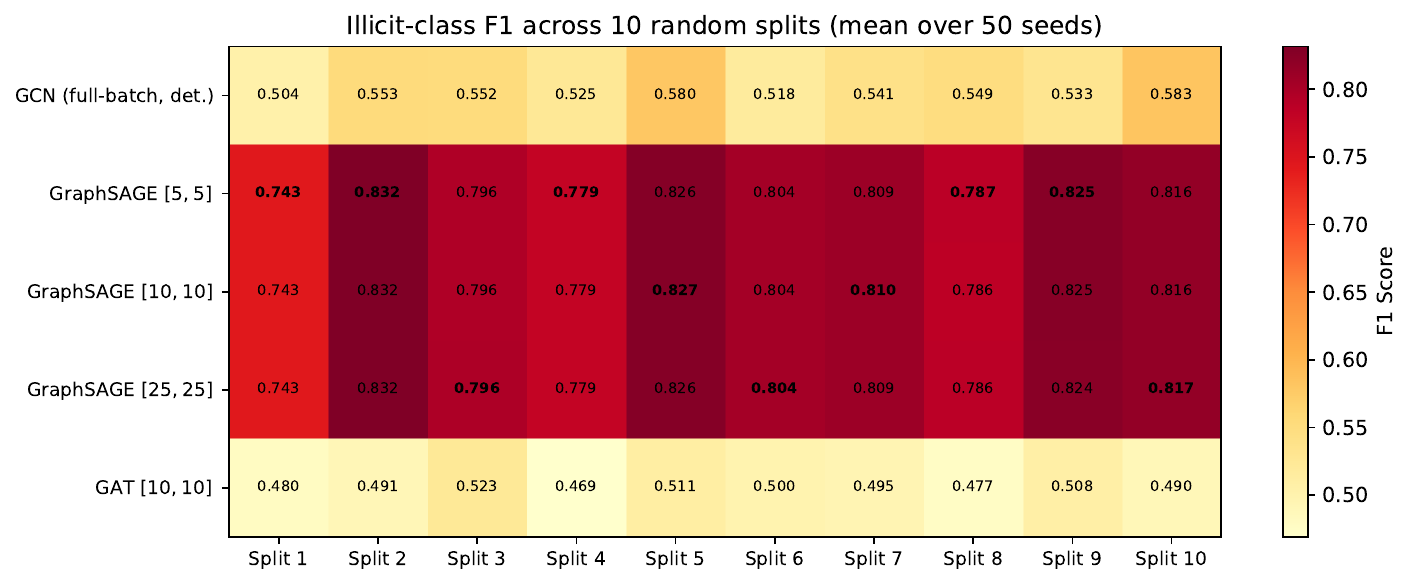}
\caption{Model ranking across 10 splits on Elliptic Bitcoin. GraphSAGE has the highest
mean F1 on \emph{every} split, so the architecture does dominate; the caption in
v1--v2 stated the opposite. The instability is one level down: which of three
replicate runs of the same GraphSAGE configuration wins a given split varies, at
margins of $1\times10^{-5}$ to $7\times10^{-4}$ F1. That is the sharper result, since
selection instability survives fixing the architecture, the protocol and the seeds.}
\label{fig:gnn_ranking_heatmap}
\end{figure}

Rankings are split-dependent, though not as v1--v2 described: GraphSAGE has the highest mean F1 on \emph{all ten} splits (Figure~\ref{fig:gnn_ranking_heatmap}). What varies is which of three replicate runs of the \emph{same} GraphSAGE configuration wins a given split, at margins of $1\times10^{-5}$ to $7\times10^{-4}$ F1. Full-batch GCN requires $5\times$ more VRAM. Embedding-level analysis (Appendix~\ref{app:embedding}) shows GAT exhibits $1.4\times$ higher cosine embedding variance (formally defined as $\mathcal{D}_{\text{cos}}$ in Section~6) than GCN, while initialization-induced drift dominates over sampling noise.

\begin{table}[t]
\centering
\caption{Illicit-class F1 on Elliptic Bitcoin, 50 seeds $\times$ 10 splits, all models
full-batch. \textbf{Flip rate} is the fraction of test nodes deviating from their
cross-seed modal label at $R{=}50$, computed within each split and averaged across the
ten; it grows with $R$, so a band written against it must state its own $R$.
\textbf{EPD} is the $R$-invariant expected pairwise disagreement, which is what we
recommend bands be written against. Neither bounds mechanical nondeterminism: the
ordering here is the reverse of the score-dispersion ordering (Section~3.1).
$\downarrow$=lower is better. These values supersede those in v1--v2, which were
computed on a single split.}
\label{tab:elliptic_f1}
\small
\resizebox{\columnwidth}{!}{%
\begin{tabular}{lcccc}
\toprule
\textbf{Model} & \textbf{F1 (mean$\pm$std)} & \textbf{Flip Rate$\downarrow$} & \textbf{EPD$\downarrow$} & \textbf{VRAM} \\
\midrule
GCN (full-batch, det.) & $0.554 \pm 0.159$ & 27.5\% & 5.4\% & 5.0\,GB \\
GraphSAGE (full-batch) & $0.799 \pm 0.127$ & 15.5\% & 1.7\% & 1.0\,GB \\
GAT (full-batch) & $0.500 \pm 0.117$ & 25.7\% & 6.5\% & 1.3\,GB \\
\bottomrule
\end{tabular}}
\end{table}

\subsection{Temporal Asynchrony and Dynamic Graph Instability}

Beyond spatial sampling variance, AML networks suffer from severe temporal nondeterminism. Financial transaction networks are Continuous-Time Dynamic Graphs (CTDGs) where edges stream asynchronously and continuously \cite{rossi2020temporal}. When a transaction occurs, the state of the participating nodes must be instantaneously updated.

In live production systems relying on Temporal Graph Networks (TGNs)~\cite{rossi2020temporal}, message-passing mechanisms depend strictly on the chronological ordering of edges. TGNs maintain a node memory state $\mathbf{s}_v(t)$, which is updated via a Recurrent Neural Network (RNN) module (e.g., a GRU) based on the previous state and incoming messages:
\[ \mathbf{s}_v(t) = \text{GRU} \left( \mathbf{s}_v(t^-), \mathbf{m}_v(t) \right) \]
High-throughput distributed architectures frequently experience microsecond variances in network latency. Consequently, the exact temporal order in which simultaneous transactions are ingested and processed by the inference server fluctuates. Because the memory update function is stateful and highly path-dependent, a microsecond race condition between two edges altering the same target node fundamentally shifts the trajectory of the localized embedding, breaking exact replication between offline backtests and live deployment. While no published study has yet quantified the exact magnitude of temporal-ordering variance in production AML graphs, the analogous finding from LLM inference---where batch-ordering changes alone induce $>$4\% accuracy variance \cite{ding2025deterministic}---suggests that path-dependent financial graph systems face similar or greater sensitivity.

\subsection{Hardware-Level Nondeterminism in Scatter Operations}

Even if a financial institution completely freezes the graph topology and strictly sequences edge ingestion, exact computational reproducibility remains elusive. In frameworks commonly used for GNNs (such as PyTorch Geometric), the $\text{AGGREGATE}$ function relies on scatter-reduction operations (e.g., \texttt{scatter\_add}) to pool messages from a variable number of source nodes. On heavily parallelized GPU architectures, these scatter operations are implemented using atomic hardware instructions (e.g., \texttt{atomicAdd} in CUDA) to prevent memory corruption across parallel threads \cite{fey2019fast}.

Floating-point arithmetic in limited precision is inherently non-associative; meaning $(\mathbf{a} + \mathbf{b}) + \mathbf{c} \neq \mathbf{a} + (\mathbf{b} + \mathbf{c})$. Because the GPU warp scheduler dynamically dictates the execution order of parallel threads based on instantaneous hardware load, the exact order in which neighbor messages are accumulated is strictly nondeterministic across independent runs, posing a direct challenge to verifiable AI systems.

\subsection{Mitigations for Graph Nondeterminism}

Three approaches address GNN reproducibility at different cost points. First, \texttt{torch.use\_deterministic\_algorithms(True)} forces deterministic scatter operations at the framework level, but requires full-batch processing ($5\times$ VRAM, as shown in Table~\ref{tab:elliptic_f1}). Second, increasing the neighbor sample size reduces sampling variance with diminishing returns~\cite{hamilton2017graphsage}, trading memory for stability; v1--v2 attributed this to ``our experiments'', which was incorrect, since no neighbor sampling is exercised anywhere in our experiment. Third, for temporal graphs, strictly sequencing edge ingestion via timestamp-ordered queues eliminates race conditions at the cost of throughput. Unlike the tabular setting where exact solutions exist (TreeSHAP), no current approach achieves zero-cost determinism for large-scale GNNs---the memory-determinism trade-off remains fundamental.

\section{Agentic AML Workflows and Unstructured Nondeterminism}

While tabular and graph-based risk models process numerical data, the vast majority of financial compliance relies on parsing unstructured, multimodal text. Historically, institutions relied on deterministic rule engines or static embedding models (e.g., FinBERT) for Named Entity Recognition (NER) \cite{huang2022finbert}. Because these architectures rely on a single forward pass without autoregressive generation, they achieve near-perfect computational reproducibility.

However, the industry has rapidly shifted toward deploying Large Language Models (LLMs) and multi-agent systems. These systems actively reason, execute tool calls, and synthesize multi-document risk narratives. The trend toward general-purpose reasoning models trained with reinforcement learning~\cite{wang2025nemotron} further amplifies this concern: extended chain-of-thought sequences generate hundreds or thousands of tokens, providing more opportunities for micro-variances to compound. This transition from static embeddings to autoregressive, agentic reasoning introduces a significant vulnerability: the breakdown of exact-match computational determinism.

\subsection{Failure of Greedy Decoding}

Setting the generation temperature to zero ($T{=}0$) for greedy decoding does not guarantee determinism in practice~\cite{wang2025assessing}. Zhang et al.~\cite{ding2025deterministic} show changing tensor-parallel (TP) size induces $>$4\% accuracy variance even under greedy decoding; their TBIK kernels restore bit-wise reproducibility. Inference determinism is a \emph{systems} property~\cite{crossdrift2025}. In Chain-of-Thought (CoT) pipelines, a single token substitution at step $t$ irreversibly alters the context for $t{+}1$, compounding into divergent risk summaries. Hardware-level nondeterminism further compounds this: optimized kernels (FlashAttention~\cite{dao2022flashattention}) rely on atomic reductions whose accumulation order varies with batch size~\cite{yuan2025mitigating}.

\paragraph{Empirical validation.}
We process 100 synthetic SAR narratives with three LLMs (Qwen2.5-7B, InternLM2.5-7B, Phi-3-mini) via vLLM on A10G, greedy decoding, varying only TP$\in\{1,2,4\}$ (10 runs/config). Table~\ref{tab:llm_determinism} reports results using four metrics: Exact Match (EM, binary string identity), Pairwise Semantic Determinism (PSD, sentence-level semantic equivalence; see Section~6), Entity-level Jaccard (Ent.J), and the Token Determinism Index. \textbf{The columns marked $\text{TDI}^p$ and $\text{PSD}^p$ are proxies, not the metrics as defined in Section~6}, and are discussed as such below; the definitions require per-token logprobs and a sentence-embedding model respectively, and the recorded runs retained neither.

Within any fixed TP, every run is bit-identical and EM${=}$1.0, so the pipeline appears perfectly deterministic to any test that holds infrastructure constant. Across TP widths, \textbf{$30$--$36\%$ of prompts yield different text}. \emph{Versions v1--v2 reported ``14--18\% diverge'' here, which is exactly half the true rate.} That figure came from pooling both widths and reading $1-\text{EM}$ as the divergence rate: because each configuration is internally bit-identical, per-prompt EM takes only the values $0.5$ and $1.0$, so the pooled mean is $1-d/2$ for a divergence fraction $d$, and $1-\text{EM}$ recovers $d/2$. We report $d$ directly; a metric with a floor at half its range is a poor compliance threshold. Entity-level Jaccard remains high ($0.96$--$0.99$), so most surface divergence leaves extracted fields intact --- but the residual gap below $1.0$ is the compliance-relevant part, since a $1$--$4\%$ chance that an entity or amount differs disqualifies a filed field even when the narrative reads the same.

\begin{table}[t]
\centering
\caption{LLM determinism on SAR extraction (100 prompts, 10 runs/config, $T{=}0$).
$\uparrow$=higher is better, $\downarrow$=lower is better. $d$ is the fraction of
prompts whose text differs across the compared widths, computed directly rather than as
$1-\text{EM}$: pooled EM has a floor at $0.5$ because each configuration is internally
bit-identical, so v1--v2's reading of $1-\text{EM}$ as the divergence rate reported half
its true value. $\text{PSD}^p$ and $\text{TDI}^p$ are the \emph{proxy} implementations
used in v1--v2, retained for comparison; \textbf{PSD mean} and \textbf{PSD min} are the
metric as defined, recomputed from SemScore over the cached generations. Note that the
proxy's apparent ``substantive drift'' for InternLM ($0.88$) does not survive: its
faithful PSD is $0.996$. Phi-3's faithful \emph{minimum} of $0.598$ is the figure a
tolerance on the worst narrative would have to clear.}
\label{tab:llm_determinism}
\small
\resizebox{\columnwidth}{!}{%
\begin{tabular}{llccccccc}
\toprule
\textbf{Model} & \textbf{Comparison} & \textbf{EM$\uparrow$} & \textbf{$d\downarrow$} & \textbf{PSD$^p\uparrow$} & \textbf{PSD mean$\uparrow$} & \textbf{PSD min$\uparrow$} & \textbf{Ent.J$\uparrow$} & \textbf{TDI$^p\uparrow$} \\
\midrule
\multirow{2}{*}{Qwen2.5-7B}
 & Within (any TP) & 1.00 & 0\% & 1.00 & 1.000 & 1.000 & 1.00 & 1.00 \\
 & TP1 vs.\ 4 & 0.85 & 30\% & 0.99 & 0.997 & 0.923 & 0.99 & 0.77 \\
\addlinespace
\multirow{2}{*}{InternLM2.5}
 & Within (any TP) & 1.00 & 0\% & 1.00 & 1.000 & 1.000 & 1.00 & 1.00 \\
 & TP1 vs.\ 4 & 0.82 & 36\% & 0.88 & 0.996 & 0.975 & 0.97 & 0.75 \\
\addlinespace
\multirow{2}{*}{Phi-3-mini}
 & Within (any TP) & 1.00 & 0\% & 1.00 & 1.000 & 1.000 & 1.00 & 1.00 \\
 & TP1 vs.\ 4 & 0.83 & 35\% & 0.89 & 0.992 & \textbf{0.598} & 0.96 & 0.83 \\
\bottomrule
\end{tabular}}
\end{table}



\subsection{Trajectory Drift and Agentic Nondeterminism}

When LLMs are deployed as agents (ReAct~\cite{yao2022react}), nondeterminism compounds through tool-call chains. DFAH~\cite{dfah2025} finds 120B+ models require $3.7\times$ larger validation samples than 7--20B models. Wang et al.~\cite{wang2025sted} report 19--46\% structured-output degradation across temperatures. Lee et al.~\cite{lee2026aema} show single-judge MAE is $4\times$ higher than multi-agent evaluation on identical financial inputs (see Table~\ref{tab:llm_instability} in Appendix~\ref{app:llm_evidence} for a full evidence summary).

\section{Engineering Determinism: Architectural Gateways and Systems Solutions}

Resolving the reproducibility crisis in financial risk requires acknowledging that monolithic, probabilistic deep learning models cannot simultaneously achieve high-level reasoning and strict mathematical determinism. Consequently, the systems engineering community is pivoting toward hybrid architectures that isolate stochastic generation from deterministic execution.

\subsection{Neuro-Symbolic Gateways and Strict API Tooling}

To combat agentic trajectory drift, enterprise systems are moving toward strictly typed, neuro-symbolic architectures. In these frameworks, the LLM is restricted to a semantic translation layer---reading unstructured text and mapping it to a predefined ontology. All mathematical calculations, logical deductions, and final risk scoring are offloaded to a deterministic symbolic engine. By enforcing strict API contracts (e.g., forcing JSON schema adherence via constrained decoding frameworks), institutions can ensure that the risk engine processes deterministic code \cite{willard2023efficient}. A recent benchmark of constrained decoding methods confirms that JSON Schema-based generation can guarantee structural compliance at the token level, though the effectiveness varies across schema complexity and model size \cite{tam2025structured}. The IBM cross-provider validation framework further demonstrates that combining schema enforcement with multi-provider consensus checking reduces output drift in financial reconciliation and regulatory reporting workflows \cite{crossdrift2025}.

\subsection{Batch-Invariant Inference Servers}

Addressing hardware-level nondeterminism requires intervention at the kernel level. Recent work has identified that the primary source of LLM inference nondeterminism is batch-size-dependent reduction order in GPU kernels \cite{thinkingmachines2025}. Proposed mitigations include batch-invariant kernel designs that rewrite atomic reduction operations (specifically RMSNorm and Attention accumulations) to force a fixed summation order regardless of batch composition, as well as verified speculation approaches that decouple determinism from kernel design entirely \cite{chen2025verifiedspec}. By guaranteeing that floating-point partial sums are accumulated in a strictly identical order regardless of surrounding batch size or thread scheduling, these systems can achieve run-to-run bitwise exactness, enabling standard software unit-testing to be applied to locally deployed LLMs. However, these approaches remain early-stage: batch-invariant kernels incur throughput overhead, and verified speculation requires additional engineering complexity.

\subsection{Cryptographic Audit Trails for Model Risk Management}

To satisfy regulators enforcing the EU AI Act and SR 11-7 \cite{eu2024aiact, bis2024mrm}, systems engineers are implementing cryptographic audit trails. Because API drift and model weight updates are inevitable, true reproducibility requires state-rebuilding capabilities. Modern frameworks employ hash-chain-backed audit logging, creating immutable snapshots of the exact model weights, hyperparameter states, quantization thresholds, and random seed initializations at the exact millisecond a credit denial or AML alert is generated. The DFAH framework demonstrates one practical instantiation of this approach: by logging full agent trajectories with determinism scores at each step, it enables post-hoc audit replay and pinpoints exactly where in a multi-step workflow the output diverged \cite{dfah2025}. This allows auditors to reconstruct the precise computational environment after the fact, though the storage and engineering overhead of maintaining such trails at enterprise scale remains an open challenge.

\subsection{Cross-Modality Synthesis}

Table~\ref{tab:synthesis} synthesizes the reproducibility failure modes, evidence, and mitigations across all three modalities. Each domain faces distinct mechanical sources of nondeterminism, but the regulatory implication is shared: systems that cannot demonstrate reproducible outputs under controlled conditions cannot satisfy audit requirements regardless of predictive accuracy.

\begin{table*}[t]
\centering
\caption{Cross-modality synthesis of reproducibility failures in financial AI. $^\dagger$First-party results.}
\label{tab:synthesis}
\small
\begin{tabular}{p{1.3cm}p{2.0cm}p{3.8cm}p{2.8cm}p{3.2cm}}
\toprule
\textbf{Domain} & \textbf{Failure Mode} & \textbf{Evidence} & \textbf{Mitigation} & \textbf{Trade-off} \\
\midrule
Tabular & Sampling variance & 17,437$\times$ divergence~\cite{covert2021improving}; $J@3{=}0.71$--$0.76$$^\dagger$ & TreeSHAP; paired sampling & Computation cost \\
Tabular & Adversarial fragility & 85--100\% attack success~\cite{slack2020fooling} & Adversarial regularization~\cite{ballegeer2025creditstability} & Testing burden \\
Graph & Split/init sensitivity & Rankings flip across splits~\cite{shchur2018pitfalls}; 3/10$^\dagger$ & Repeated-run evaluation & Benchmarking cost \\
Graph & Stochastic sampling & 30\% flip rate on Elliptic$^\dagger$ & Deterministic aggregation & $5\times$ VRAM$^\dagger$ \\
LLM & TP/batch config & $>$4\% accuracy var.~\cite{ding2025deterministic}; EM$=$0.82--0.85$^\dagger$ & Batch-invariant kernels & Throughput overhead \\
LLM & Trajectory drift & MAE $4\times$ higher~\cite{lee2026aema}; 19--46\% degrad.~\cite{wang2025sted} & Process-aware eval. & Orchestration cost \\
\bottomrule
\end{tabular}
\end{table*}

\section{Quantifying Reproducibility: Evaluation Metrics and Frameworks}

Traditional machine learning benchmarks overwhelmingly prioritize predictive accuracy metrics---such as Area Under the Receiver Operating Characteristic Curve (AUROC), F1-score, and Mean Average Precision (mAP). However, these metrics evaluate the relationship between the model's output and the ground-truth label, completely ignoring the mechanical stability of the model across independent execution traces. While domain-agnostic reproducibility tools exist---notably TorchMetrics \cite{detlefsen2022torchmetrics}, which provides standardized metric computation for reproducible evaluation, and Bhojanapalli et al. \cite{bhojanapalli2021reproducibility}, who quantify prediction-level variance across hardware configurations---these frameworks do not address the modality-specific failure modes that arise in regulated financial settings: stochastic feature attribution in credit scoring, topological embedding drift in fraud detection graphs, and trajectory divergence in agentic AML workflows. To rigorously audit financial AI systems for regulatory compliance, the machine learning community must adopt a standardized taxonomy of reproducibility metrics that quantify internal variance independent of ground-truth accuracy, tailored to the specific computational modalities deployed in financial risk management.

The measurement of determinism requires modality-specific evaluation frameworks. To accommodate this, we propose a layered evaluation stack that transitions from continuous probability scoring (DDR) to tabular feature ranking (RBO), latent graph embeddings (Cosine Variance), and unstructured agentic generation (TDI and PSD).

\subsection{Continuous Determinism (DDR)}

For continuous prediction scores, Anjum et al.~\cite{anjum2024ddr} propose the Deterministic-Non-Deterministic Ratio: $\text{DDR} = \text{Var}(D)/\text{Var}(N)$, decomposing predictions into signal $D$ and hardware noise $N$. We include DDR in our taxonomy for completeness but do not validate it experimentally (see Appendix~\ref{app:ddr} for details).

\subsection{Feature Attribution Stability in Tabular Models}

In credit risk, where post-hoc explainers like KernelSHAP are used to generate adverse action notices, reproducibility is measured by the stability of the generated feature importance rankings across multiple runs with distinct random seeds. Given a tabular model $f$ and an input instance $x$, an explainer generates a ranked list of feature attributions $R$. 

To quantify the deterministic stability of the explainer, practitioners cannot rely on simple Pearson correlation, as regulatory compliance primarily cares about the top reasons for denial, not the bottom. Therefore, the standard measurement is Top-$k$ Feature Agreement, frequently calculated using the Jaccard Index or Rank-Biased Overlap (RBO) \cite{webber2010similarity, ghorbani2019interpretation}. 

For two independent runs yielding top-$k$ feature sets $S_1$ and $S_2$, the Jaccard Feature Intersection is defined as:
\[ J@k(S_1, S_2) = \frac{|S_1 \cap S_2|}{|S_1 \cup S_2|} \]
A deterministic, legally compliant system requires $J@k = 1.0$ for small $k$ (e.g., the top 4 denial reasons). As the evidence in Section~2 demonstrates, sampling-based explainers can produce substantial rank disagreement in mid-tier features, implying $J@k < 1.0$ for practically relevant values of $k$. To penalize rank inversions at the top of the list more heavily, RBO calculates the expected overlap at varying depths $d$, with a decay parameter $p$:
\[ \text{RBO}(R_1, R_2, p) = (1 - p) \sum_{d=1}^{\infty} p^{d-1} \frac{|R_{1:d} \cap R_{2:d}|}{d} \]
High-variance models exhibit low RBO scores, providing a quantitative metric to disqualify unstable credit models from production deployment.

\subsection{Latent Space Variance in Graph Networks}

For Graph Neural Networks (GNNs) used in transaction monitoring, the final classification (e.g., fraud vs. benign) often remains stable, but the underlying latent representations fluctuate wildly due to stochastic neighborhood sampling and asynchronous edge ingestion. Measuring GNN reproducibility requires quantifying embedding drift in the high-dimensional latent space.

Given a target node $v$, let $\mathbf{h}_v^{(s_1)}$ and $\mathbf{h}_v^{(s_2)}$ represent the final layer $L_2$-normalized embeddings generated across two independent runs with distinct sampling seeds $s_1$ and $s_2$. The reproducibility of the GNN is quantified by the expected pairwise Cosine Distance across $N$ runs:
\[ \mathcal{D}_{\text{cos}}(v) = \frac{2}{N(N-1)} \sum_{i=1}^{N} \sum_{j=i+1}^{N} \left( 1 - \frac{\mathbf{h}_v^{(s_i)} \cdot \mathbf{h}_v^{(s_j)}}{\|\mathbf{h}_v^{(s_i)}\| \|\mathbf{h}_v^{(s_j)}\|} \right) \]
In a perfectly deterministic system, $\mathcal{D}_{\text{cos}} = 0$. In production AML networks using GraphSAGE, increasing the stochastic sample size $\mathcal{S}_{\mathcal{N}(v)}$ decreases latent variance at the cost of exponentially higher VRAM consumption, formally defining the hardware-determinism trade-off \cite{hamilton2017graphsage}.

\paragraph{Embedding variance vs.\ decision stability.} High $\mathcal{D}_{\text{cos}}$ does not necessarily imply unstable decisions: relative scores (e.g., $\mathbf{h}_v \cdot \mathbf{h}_u$) may remain stable if embeddings drift coherently. Our prediction flip rate (Section~3) directly measures decision-level instability and should be used alongside $\mathcal{D}_{\text{cos}}$.

\subsection{Trajectory Determinism in Agentic Workflows}

For multi-agent AML pipelines, determinism must be measured over the action path itself. An agent's trace can be modeled as a sequence of discrete actions; applying the Levenshtein distance across independent executions pinpoints where the trajectory drifted~\cite{dfah2025}. However, action-path metrics and Exact Match share a limitation: they cannot distinguish genuine uncertainty from legitimate variability among equally valid outputs. The following subsections introduce complementary metrics addressing this gap.

\subsection{Logit-Level Determinism Scoring}

The metrics discussed above---Exact Match and Trajectory Edit Distance---as well as token-level semantic overlap measures like BERTScore~\cite{zhang2019bertscore}, all operate on the \emph{output} side of the inference pipeline: they compare final generated tokens or action sequences across runs. However, a complementary and arguably more fundamental approach is to measure determinism at the \emph{logit} level, directly inspecting the pre-softmax score distribution that governs token selection. Recent work on Logits-induced Token Uncertainty (LogU) provides a principled framework for this analysis \cite{chen2025logu}.

The key insight of LogU is that probability-based uncertainty measures lose critical information during softmax normalization. When the raw logit vector $\mathbf{z}_t = \{\mathcal{M}(\tau^m | \mathbf{q}, \mathbf{a}_{t-1})\}_{m=1}^{|\mathcal{V}|}$ is normalized into a probability distribution, the \emph{absolute magnitude} of evidence accumulated during training is discarded, retaining only relative token rankings. This creates a fundamental ambiguity: a token with probability 0.3 could reflect either genuine model uncertainty (the model lacks knowledge and distributes mass thinly) or confident multi-answer awareness (the model has encountered many valid continuations and distributes mass across several strong candidates). For financial determinism auditing, this distinction is critical---the former signals unreliable inference, while the latter signals robust but legitimately variable output.

LogU resolves this ambiguity by treating the top-$K$ logits as parameters of a Dirichlet distribution $\text{Dir}(\alpha_1, \ldots, \alpha_K)$, where $\alpha_k = \mathcal{M}(\tau_k | \mathbf{q}, \mathbf{a}_{t-1})$ for the $k$-th highest-scoring token. This decomposition yields two orthogonal uncertainty dimensions. The \textit{aleatoric uncertainty} (AU) captures the relative entropy among candidate tokens:
\[ \text{AU}(a_t) = -\sum_{k=1}^{K} \frac{\alpha_k}{\alpha_0} \left( \psi(\alpha_k + 1) - \psi(\alpha_0 + 1) \right), \quad \alpha_0 = \sum_{k=1}^{K} \alpha_k \]
where $\psi$ denotes the digamma function. The \textit{epistemic uncertainty} (EU) captures the model's overall evidence strength:
\[ \text{EU}(a_t) = \frac{K}{\sum_{k=1}^{K} (\alpha_k + 1)} \]

We operationalize this as the \textit{Token Determinism Index} (TDI). Given a sequence of $T$ generated tokens, TDI measures the fraction generated with sufficient epistemic evidence:
\[ \text{TDI} = \frac{1}{T} \sum_{t=1}^{T} \mathds{1}\left[\text{EU}(a_t) < \theta_{\text{EU}}\right] \]
where $\theta_{\text{EU}}$ is a calibrated threshold. A TDI close to 1.0 indicates nearly all tokens were generated with sufficient evidence strength, regardless of whether the model chose among multiple valid candidates. Unlike Exact Match, TDI tolerates legitimate variability while flagging genuinely unreliable tokens.

\subsection{Semantic Similarity Scoring for Output Equivalence}

Regulators need to assess whether independent runs convey the \emph{same financial meaning}. We adapt SemScore~\cite{aynetdinov2024semscore}---which achieves the highest human-evaluation correlation among 8 metrics---into \textit{Pairwise Semantic Determinism} (PSD). Given $N$ runs producing outputs $O_i$:
\[ \text{PSD} = \frac{2}{N(N-1)} \sum_{i=1}^{N} \sum_{j=i+1}^{N} \text{SemScore}(O_i, O_j) \]
PSD is computable without model internals, complementing token-level BERTScore~\cite{zhang2019bertscore}.

\paragraph{Limitations.} Sentence-level PSD can be dominated by filler words, masking divergences in critical content (entity names, amounts). For structured outputs, a matching-based approach---extracting information units and solving an optimal assignment (e.g., Hungarian algorithm)---provides a more precise signal; our entity-level Jaccard approximates this (see Appendix~\ref{app:psd_limitations} for a full discussion).

Together, TDI and PSD answer: (1)~did the model have sufficient evidence? and (2)~do runs convey the same meaning? Table~\ref{tab:determinism_metrics} summarizes the full metric taxonomy.

\begin{table}[t]
\centering
\caption{Proposed determinism metric taxonomy for financial AI.}
\label{tab:determinism_metrics}
\small
\resizebox{\columnwidth}{!}{%
\begin{tabular}{p{1.8cm}p{1.4cm}p{2.2cm}p{1.4cm}p{2.5cm}}
\toprule
\textbf{Metric} & \textbf{Layer} & \textbf{Measures} & \textbf{Access} & \textbf{Use Case} \\
\midrule
J@k / RBO & Feature & Attribution stability & Output & Credit adverse actions \\
$\mathcal{D}_{\text{cos}}$ & Embedding & Graph latent drift & Output & Fraud/AML monitoring \\
TDI & Logit & Per-token evidence & Internal & Diagnosing divergence \\
PSD & Sentence & Semantic equivalence & Output & Black-box API audit \\
DDR & Prediction & Signal-to-noise & Output & Classification fragility \\
\bottomrule
\end{tabular}}
\end{table}

\subsection{Empirical Validation of TDI and PSD}

\emph{This subsection reports a negative result that replaces the positive one claimed
in v1--v2.} Those versions read three regimes off a scatter of TDI against PSD and
presented them as validating the complementarity of the two measures. That reading does
not survive, for two reasons that compound.

First, the quantity plotted as PSD was not PSD. It was pairwise entity-set Jaccard,
thresholded---computed from the same entity sets the Ent.J column of
Table~\ref{tab:llm_determinism} is built from. PSD and Ent.J were therefore not two
measures but one, which is why they move together in that table. A complementarity
claim cannot rest on two views of the same quantity.

Second, the comparison was made at the wrong unit. Correlating nine configuration
means has no power to answer whether two instruments carry different information about
a given output; the question is per prompt. Recomputing PSD as defined---SemScore over
the cached generations---and correlating per prompt against entity Jaccard gives
$+0.18$ to $+0.72$ (mean $+0.49$, significant on four of six configurations), against
$+0.93$ to $+1.00$ for the proxy, which reaches exactly $1.000$ on Qwen.

What the faithful metric supports is weaker and narrower than complementarity. PSD is
\emph{partly} redundant with the entity check and not separable from it: it carries
some information the entity sets do not, and the two should be read together rather
than treated as independent evidence. It does grade where the proxy could not, taking
27--36 distinct values per 100 prompts against the proxy's two. What it does not
recover is separation between models---all three sit between $0.992$ and $0.997$ on the
mean---so the three regimes were a property of the proxy and not of the models.
Per-prompt \emph{minima} do carry signal the mean conceals (Phi-3 reaches $0.598$ on
its worst prompt against Qwen's $0.923$), which argues for reporting PSD as a minimum
when the claim concerns the worst output an analyst will read.

\begin{figure}[t]
\centering
\includegraphics[width=\columnwidth]{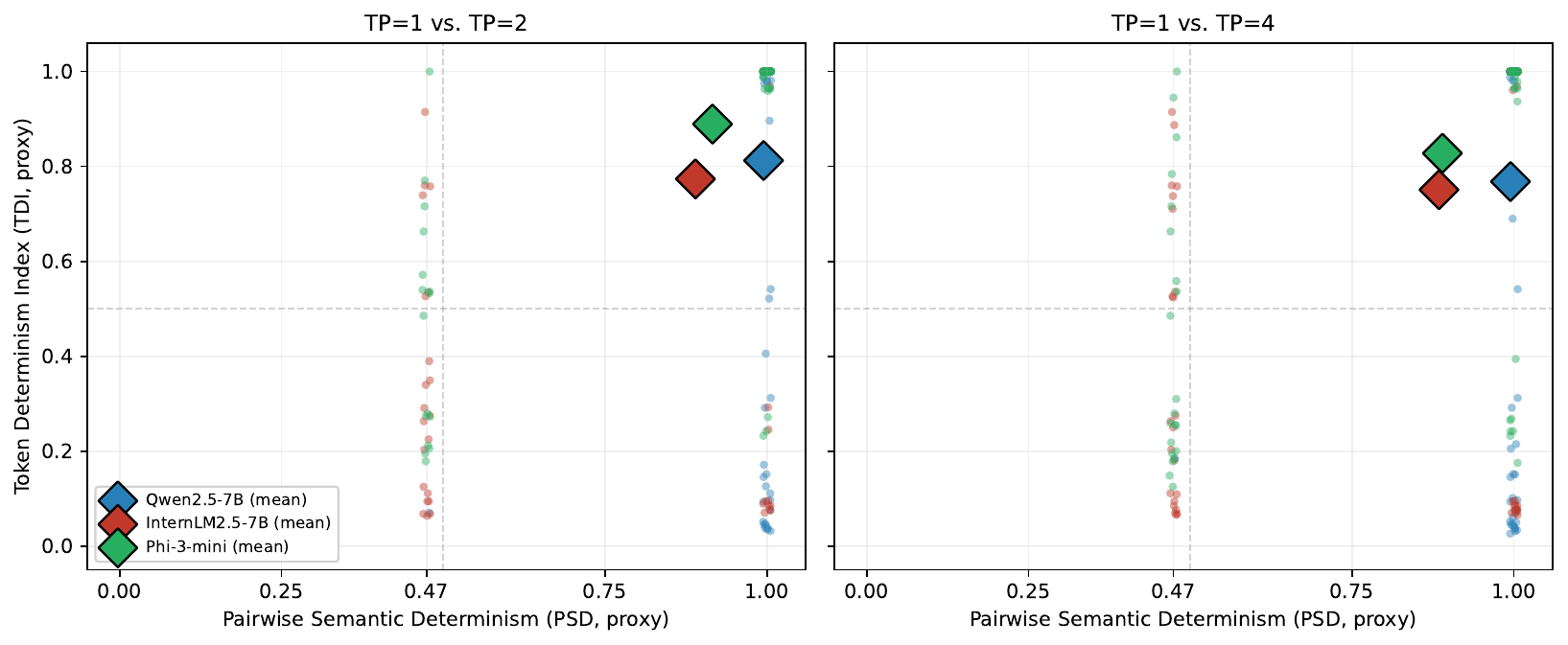}
\caption{$\text{TDI}^p$ against $\text{PSD}^p$ for all model$\times$TP configurations.
\textbf{This figure is retained as a record of a withdrawn claim.} Versions v1--v2 read
three regimes off it and presented them as showing that the two instruments capture
orthogonal aspects of determinism. They do not: the quantity plotted as PSD is a
thresholded overlap of the same entity sets the Ent.J column is built from, so the two
axes are not independent, and the comparison was made between configuration means
where the question is per prompt. This version plots the per-prompt values (small
markers) with the configuration means as diamonds, which makes the failure visible
directly: the proxy $\text{PSD}^p$ occupies essentially two values, ${\approx}0.47$
and $1.00$, so it is an indicator of whether the two widths agreed rather than a graded
semantic distance. Under the faithful metric the regimes dissolve and all three models
fall between $0.992$ and $0.997$ (Section~6.7).}
\label{fig:tdi_psd}
\end{figure}

\section{Conclusion and Future Directions}

The transition from econometric models to deep networks and multi-agent workflows has transformed the reproducibility crisis from a statistical phenomenon into a systems engineering bottleneck. In tabular credit scoring, NP-hard Shapley computation forces stochastic approximations that destabilize explainability. In graph-based fraud detection, memory bottlenecks necessitate stochastic sampling that breaks embedding replication. In agentic workflows, dynamic batching and non-associative reductions compound through autoregressive trajectories, destroying exact-match determinism.

Our first-party experiments quantify these failures on public financial datasets: KernelSHAP explanation rankings vary for roughly one applicant in four at the $n{=}100$ default ($J@3{=}0.74$ at $M{=}20$, falling to $0.61$ on the same data under a wider encoding), GNN prediction labels deviate from their cross-seed mode for $15$--$28\%$ of test nodes ($2$--$7\%$ as pairwise disagreement), and LLM outputs diverge for $30$--$36\%$ of prompts when tensor-parallel configuration changes---even under greedy decoding. The graph and LLM figures correct v1--v2, which reported the former on a single split and the latter at half its true value.

Future work must prioritize: (1)~batch-invariant CUDA kernels for bitwise accumulation without throughput loss; (2)~neuro-symbolic architectures routing LLM outputs to deterministic engines; (3)~deterministic attribution methods beyond stochastic sampling; and (4)~instruments that genuinely separate benign variability from unreliable inference. Our own attempt does not: the faithful PSD is partly redundant with an entity-level check rather than complementary to it (Section~6.7), and the faithful TDI remains unmeasured because it requires logprobs the serving layer must be configured in advance to retain. Establishing separability, rather than assuming it, is the open problem.

\paragraph{Limitations.} The temporal-ordering analysis (Section~3) describes plausible mechanisms without production measurements. Our GNN experiment uses a single public dataset (Elliptic Bitcoin) and may not generalize to proprietary enterprise-scale graphs. Our LLM experiment tests only 7B-class open-source models on synthetic prompts---proprietary models and real compliance documents may exhibit different divergence patterns. The $\text{TDI}^p$ and $\text{PSD}^p$ columns are proxies rather than the metrics defined in Section~6: $\text{TDI}^p$ is a positional token-agreement rate rather than a Dirichlet decomposition over stored logits, and $\text{PSD}^p$ is a thresholded entity-set overlap rather than a semantic score. The faithful PSD is recoverable from the cached generations and is reported in Section~6.7; the faithful TDI is not, because the recorded runs never requested logprobs, and obtaining it requires re-generation with \texttt{logprobs} enabled. An instrument that consumes internal state must be provisioned before the run it is meant to audit.

\appendix

\section{LogU Uncertainty Decomposition}
\label{app:logu}

LogU~\cite{chen2025logu} treats the top-$K$ logits as Dirichlet parameters $\text{Dir}(\alpha_1, \ldots, \alpha_K)$. The aleatoric uncertainty (AU) captures relative entropy:
$\text{AU}(a_t) = -\sum_{k=1}^{K} \frac{\alpha_k}{\alpha_0} ( \psi(\alpha_k + 1) - \psi(\alpha_0 + 1) )$, $\alpha_0 = \sum_k \alpha_k$.
The epistemic uncertainty (EU) captures evidence strength:
$\text{EU}(a_t) = K / \sum_{k=1}^{K} (\alpha_k + 1)$.
This yields four quadrants: (I)~high AU+EU = unreliable; (II)~low AU, high EU = best guess; (III)~low AU+EU = ideal deterministic; (IV)~high AU, low EU = legitimate variability. On TruthfulQA, LogU achieves 72--74\% AUROC~\cite{chen2025logu}.

\section{Tabular Determinism Solutions}
\label{app:tabular_solutions}

Table~\ref{tab:tabular_solutions} summarizes the three structural solutions for enforcing determinism in tabular risk models discussed in Section~2.3.

\begin{table}[h!]
\centering\small
\caption{Determinism challenges and solutions in tabular risk modeling.}
\label{tab:tabular_solutions}
\resizebox{\columnwidth}{!}{%
\begin{tabular}{p{2.2cm}p{2.8cm}p{4.5cm}}
\toprule
\textbf{Challenge} & \textbf{Source} & \textbf{Mitigation} \\
\midrule
Sampling Variance & NP-hard Shapley & \textbf{TreeSHAP:} Exact routing \\
Explanation Drift & Perturbation fragility & \textbf{Adversarial regularization} \\
Black-Box & Post-hoc patches & \textbf{EBMs:} Variance-free lookups \\
\bottomrule
\end{tabular}}
\end{table}

\section{GNN Evaluation Instability}
\label{app:shchur}

Table~\ref{tab:graph_instability} reports GNN evaluation instability from Shchur et al.~\cite{shchur2018pitfalls}, demonstrating that model rankings reverse across random data splits on standard citation-network benchmarks.

\begin{table}[h!]
\centering\small
\caption{GNN instability from Shchur et al.~\cite{shchur2018pitfalls} (100 splits $\times$ 20 init.).}
\label{tab:graph_instability}
\resizebox{\columnwidth}{!}{%
\begin{tabular}{llcccc}
\toprule
\textbf{Dataset} & \textbf{Split} & \textbf{GCN} & \textbf{GAT} & \textbf{MoNet} & \textbf{GS-max} \\
\midrule
Cora & Planetoid & 81.9\scriptsize{$\pm$0.8} & \textbf{82.8}\scriptsize{$\pm$0.5} & 82.2\scriptsize{$\pm$0.7} & 77.4\scriptsize{$\pm$1.0} \\
Cora & Random & \textbf{79.0}\scriptsize{$\pm$0.7} & 77.9\scriptsize{$\pm$0.7} & 77.9\scriptsize{$\pm$0.7} & 74.5\scriptsize{$\pm$0.6} \\
PubMed & Planetoid & \textbf{79.0}\scriptsize{$\pm$0.5} & 77.0\scriptsize{$\pm$1.3} & 77.7\scriptsize{$\pm$0.6} & 76.6\scriptsize{$\pm$0.8} \\
PubMed & Random & 69.5\scriptsize{$\pm$1.0} & 69.5\scriptsize{$\pm$0.6} & \textbf{70.7}\scriptsize{$\pm$0.5} & 70.3\scriptsize{$\pm$0.8} \\
\bottomrule
\end{tabular}}
\end{table}

\section{Elliptic Embedding Variance}
\label{app:embedding}

Table~\ref{tab:elliptic_embedding} reports the cosine embedding variance $\mathcal{D}_{\text{cos}}$ for GNN architectures on the Elliptic Bitcoin dataset, complementing the F1 and flip rate results in Section~3.

\begin{table}[h!]
\centering\small
\caption{$\mathcal{D}_{\text{cos}}$ on Elliptic Bitcoin (1K test nodes). $\downarrow$=better.}
\label{tab:elliptic_embedding}
\resizebox{\columnwidth}{!}{%
\begin{tabular}{lcc}
\toprule
\textbf{Model} & $\mathcal{D}_{\text{cos}}\downarrow$ & \textbf{Rel.\ GCN} \\
\midrule
GCN (det.) & 0.680 & $1.0\times$ \\
GraphSAGE $[5,5]$ & 0.681 & $1.0\times$ \\
GAT $[10,10]$ & 0.952 & $1.4\times$ \\
\bottomrule
\end{tabular}}
\end{table}

\section{LLM/Agentic Evidence}
\label{app:llm_evidence}

Table~\ref{tab:llm_instability} summarizes the source-verified evidence on LLM and agentic nondeterminism discussed in Section~4.

\begin{table}[h!]
\centering\small
\caption{LLM/agentic nondeterminism evidence.}
\label{tab:llm_instability}
\resizebox{\columnwidth}{!}{%
\begin{tabular}{p{1.6cm}p{1.8cm}p{1.6cm}p{1.5cm}p{3.0cm}}
\toprule
\textbf{Study} & \textbf{Setting} & \textbf{Failure} & \textbf{Metric} & \textbf{Finding} \\
\midrule
Zhang~\cite{ding2025deterministic} & TP=1--8 & Config div. & Accuracy & $>$4\% var. \\
Wang~\cite{wang2025sted} & JSON, 6 LLMs & Drift & STED & 19--46\% degrad. \\
Lee~\cite{lee2026aema} & Invoice, 30 runs & Path var. & MAE & 0.077 vs.\ 0.018 \\
\bottomrule
\end{tabular}}
\end{table}

\section{Broader Impacts}
\label{app:broader}

The inability to deterministically reproduce AI decisions in credit scoring and fraud detection poses societal risks concerning systemic bias and ECOA violations. The solutions discussed---batch-invariant architectures and cryptographic audit trails---aim to provide mathematically verifiable proofs of decision-making.

\section{PSD Limitations and Matching-Based Alternatives}
\label{app:psd_limitations}

A known limitation of sentence-level PSD is that filler words (articles, prepositions) dominate the sentence embedding, potentially masking divergences in critical content such as entity names or monetary amounts. For structured financial outputs where key fields are identifiable, a matching-based approach may be more appropriate: extract information units (entities, amounts, dates) from each output, then solve an optimal assignment problem (e.g., via the Hungarian algorithm) to find the best alignment between two runs, scoring determinism by the fraction of matched units. Our entity-level Jaccard metric (Table~\ref{tab:llm_determinism}) approximates this approach for named entities. We retain sentence-level PSD for unstructured narratives where field extraction is not straightforward, while acknowledging that field-level matching provides a more precise signal for structured compliance outputs.

\section{DDR Metric Details}
\label{app:ddr}

The Deterministic-Non-Deterministic Ratio (DDR)~\cite{anjum2024ddr} posits that any prediction $Y = D + N$, where $D$ is the deterministic signal and $N$ is nondeterministic noise from hardware scatter operations, temporal latency, or sampling variance. DDR $= \text{Var}(D)/\text{Var}(N)$ measures classification fragility: in a low-DDR state, noise can overpower the signal near the decision boundary (e.g., a 0.51 fraud probability), causing labels to flip across runs. DDR-Accuracy curves allow institutions to stress-test how rapidly accuracy degrades as noise increases, establishing minimum-DDR thresholds for deployment. Isolating $D$ from $N$ requires controlled multi-hardware experiments beyond this survey's scope.

\end{document}